\documentclass{article}
\usepackage{spconf,amsmath,graphicx,hyperref}
\usepackage{etoolbox}
\AtBeginEnvironment{thebibliography}{\footnotesize}
\usepackage{enumitem}
\usepackage{cite}
\usepackage{amsmath}
\usepackage{amssymb}
\usepackage{booktabs}
\usepackage{multirow}
\usepackage{microtype}

\usepackage{xcolor}
\usepackage{utfsym}
\hypersetup{hidelinks} 

\title{LAKAN: LANDMARK-ASSISTED ADAPTIVE KOLMOGOROV-ARNOLD NETWORK FOR FACE FORGERY DETECTION}
\name
{Jiayao Jiang\textsuperscript{1,2}, Bin Liu\textsuperscript{1,2\dag}, Qi Chu\textsuperscript{1,2}, Nenghai Yu\textsuperscript{1,2}\thanks{\dag Corresponding author: \href{mailto:flowice@ustc.edu.cn}{flowice@ustc.edu.cn}}}

\address{ 
\textsuperscript{1}School of Cyber Science and Technology, University of Science and Technology of China \\ 
\textsuperscript{2}Anhui Province Key Laboratory of Digital Security \\
  \\
  \\
}

\begin{document}
\ninept
\maketitle
\begin{abstract}
The rapid development of deepfake generation techniques necessitates robust face forgery detection algorithms. While methods based on Convolutional Neural Networks (CNNs) and Transformers are effective, there is still room for improvement in modeling the highly complex and non-linear nature of forgery artifacts. To address this issue, we propose a novel detection method based on the Kolmogorov-Arnold Network (KAN). By replacing fixed activation functions with learnable splines, our KAN-based approach is better suited to this challenge. Furthermore, to guide the network's focus towards critical facial areas, we introduce a Landmark-assisted Adaptive Kolmogorov-Arnold Network (LAKAN) module. This module uses facial landmarks as a structural prior to dynamically generate the internal parameters of the KAN, creating an instance-specific signal that steers a general-purpose image encoder towards the most informative facial regions with artifacts. This core innovation creates a powerful combination between geometric priors and the network's learning process. Extensive experiments on multiple public datasets show that our proposed method achieves superior performance.
\end{abstract}

\section{Introduction}
\label{sec:intro}
In recent years, deep learning-driven face forgery techniques have achieved remarkable advancements, with generated images possessing highly convincing perceptual quality that poses a severe challenge to social security. Consequently, more advanced face forgery detection algorithms are urgently needed. While detection methods based on Convolutional Neural Networks (CNNs) and Transformers are the mainstream approaches and have demonstrated strong performance, their reliance on fixed activation functions may present challenges when attempting to model the exceptionally complex and non-linear function distributions that characterize forgery artifacts. These fixed functions, such as ReLU or GELU, apply a uniform transformation across all features, which might not be optimal for capturing the diverse and intricate patterns found in forged content.

\begin{figure}[htb]
  \centering
  \includegraphics[width=8.5cm]{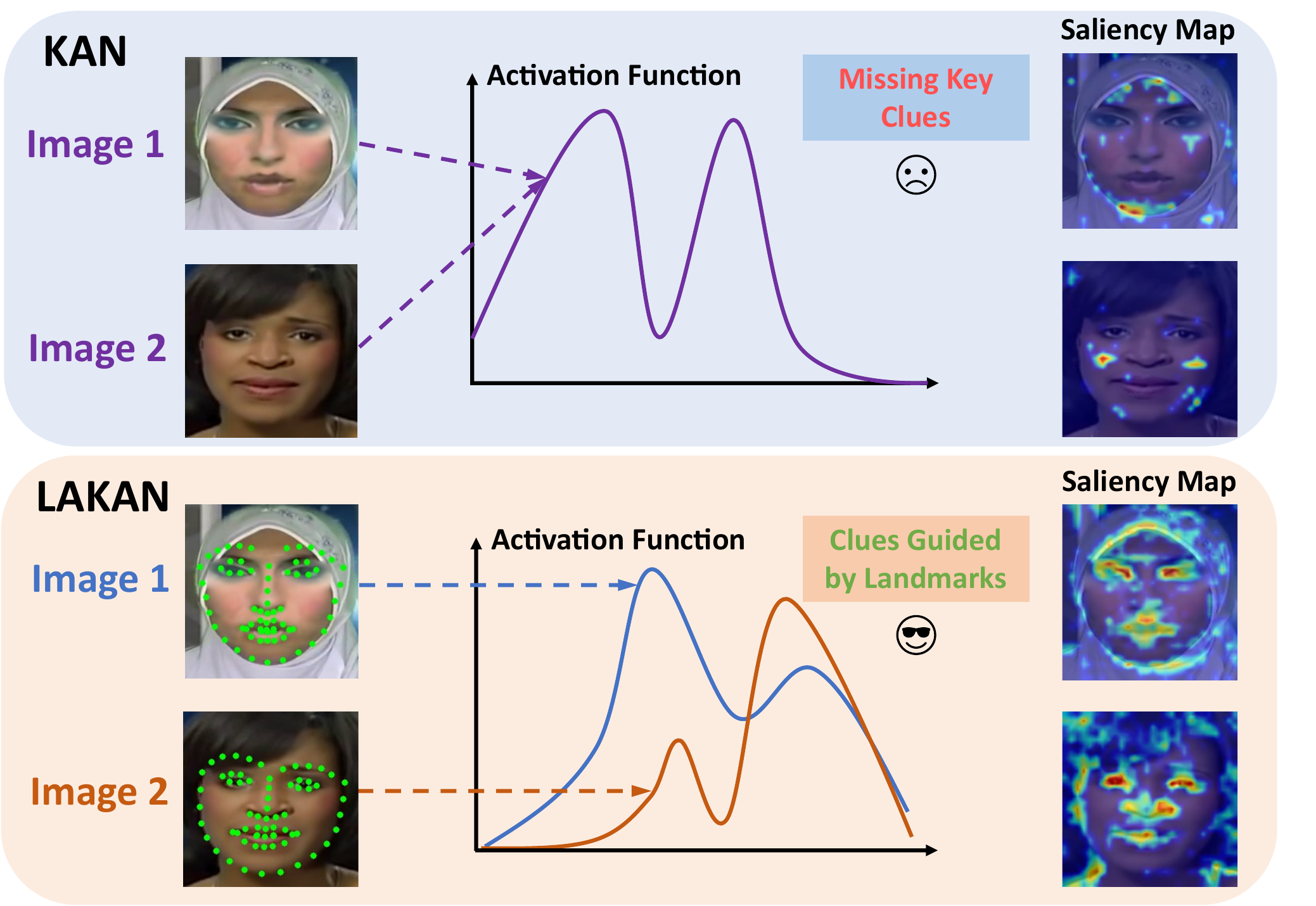}
  \vspace{-3mm}
  \caption{Compared to KAN \cite{DBLP:conf/iclr/LiuWVRHS0T25}, LAKAN adaptively constructs activation functions based on facial landmark information, enabling it to make use of the key clues provided by landmark annotations fully.}
  \label{fig:figure1}
  \vspace{-5.2mm}
\end{figure}

The recently proposed Kolmogorov-Arnold Network (KAN) \cite{DBLP:conf/iclr/LiuWVRHS0T25} offers a promising alternative. Its core innovation lies in its activation functions, which are not fixed but are represented as learnable B-spline functions on the network's edges. This design allows the splines to approximate various activation functions, granting the network superior flexibility. This intrinsic characteristic endows KAN with theoretically stronger function approximation capabilities, making it suitable for modeling the highly complex and non-linear artifacts produced by the forgery process, as it can tailor its activation functions to the specific data distribution it is learning.

However, a standard KAN applied directly to images remains a general-purpose function approximator, lacking the specific guidance needed to focus on key facial regions where forgery evidence is most likely to appear. Facial landmarks, as a powerful form of facial structure, have been shown to be crucial for forgery detection \cite{li2024landmarkbreaker}. The core idea of this paper is to leverage this prior by using facial landmarks to dynamically and adaptively construct the internal spline functions of the KAN, as illustrated in Fig. \ref{fig:figure1}. This approach allows our proposed model to generate instance-specific parameters that guide a general-purpose image encoder at different stages, steering its focus towards the most informative facial regions for each unique input. The main contributions of this paper are as follows:
\begin{figure*}[htb]
  \centering 
  \includegraphics[width=0.87\textwidth]{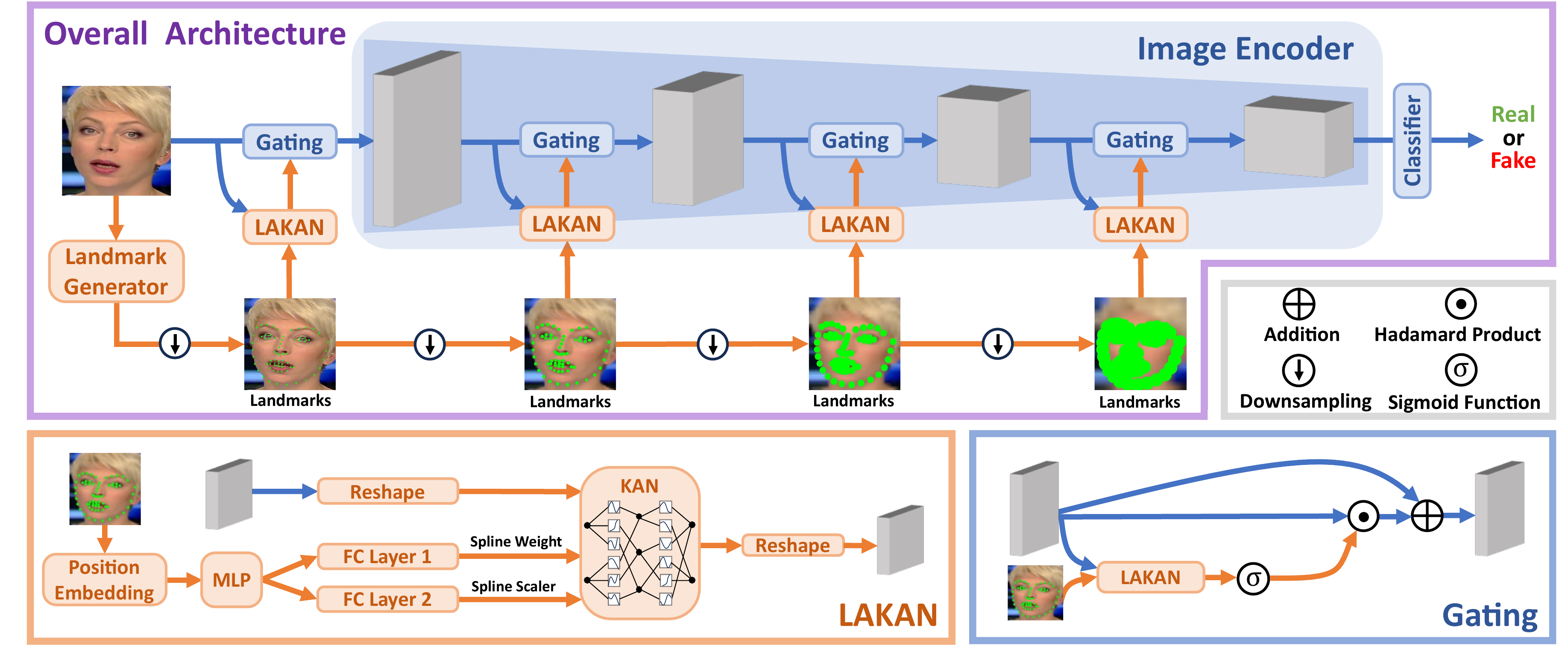}
  
  \vspace{-4mm}
  \caption{Overview of model architecture with LAKAN. The LAKAN module leverages facial landmarks to generate adaptive parameters for KAN and is applied to downsampled features from four different stages of the image encoder through gating mechanisms.}
  \label{fig:figure2}
  \vspace{-5.5mm}
\end{figure*}

\begin{itemize}[noitemsep,topsep=0pt] 
\item We introduce KAN for face forgery detection to address the challenge of capturing subtle and non-linear forgery artifacts, leveraging its function approximation capabilities.
\item We propose LAKAN, a novel module that adaptively generates KAN parameters from facial landmarks to guide an image encoder's focus towards critical facial regions.
\item Experimental results on multiple public datasets demonstrate that our proposed method achieves excellent performance.
\end{itemize}

\section{Related Work}
\label{sec:related work}

\subsection{Face Forgery Detection}
\label{ssec:subhead2.1}
The task of face forgery detection is generally treated as a two-category classification challenge. This involves a model analyzing an input, such as an image or a video, to ultimately assign it a label of either ``real'' or ``fake''. As outlined in \cite{pei2024deepfake}, the current detection methods can be organized into four categories: spatial-domain, time-domain, frequency-domain and data-driven approaches. Spatial-domain methods catch fake faces directly in the picture at image level, such as artifacts \cite{cao2022end}, saturation \cite{mccloskey2019detecting} and color \cite{he2019detection}. Time-domain approaches operate on the entire video stream, identifying manipulations by analyzing  inter-frame inconsistencies \cite{peng2024deepfakes}. Frequency-domain methods use algorithms (like FFT \cite{tan2024frequency}, DCT \cite{li2021frequency}, DWT \cite{miao2023f}) to transform data, exposing subtle forgery clues that are normally hidden. Finally, the data-driven approach is about optimizing model architectures and training them in smarter ways, extracting the maximum value from the available data \cite{yan2024transcending}.
\subsection{Kolmogorov-Arnold Network}
\label{ssec:subhead2.2}
Inspired by mathematical theorem, Kolmogorov-Arnold Network (KAN) \cite{DBLP:conf/iclr/LiuWVRHS0T25} replaces the node-based activations of MLP with learnable activation functions on the edges, enhancing both accuracy and interpretability. KAN-Conv \cite{drokin2024kolmogorov} integrates the KAN principle into convolutional layers, enhancing expressive power for vision tasks. KAT \cite{DBLP:conf/iclr/YangW25a} replaced MLP in Transformers with a parameter-efficient KAN, achieving scalable, high-performance Transformer blocks. KAN-CFD \cite{zhang2025unifying} is a novel KAN-based module for continual face forgery detection. Our proposed method LAKAN is a successful attempt and exploration in terms of general face forgery detection.

\section{METHODOLOGY}
\label{sec:methodology}
\subsection{Preliminaries}
Kolmogorov-Arnold Network \cite{DBLP:conf/iclr/LiuWVRHS0T25} is a new type of neural network that serves as an alternative to traditional MLP. Its core innovation is that it replaces fixed activation functions on nodes with learnable activation functions on edges, offering a new approach to solving the challenges of function fitting accuracy and model interpretability. It originates from the Kolmogorov-Arnold representation theorem:
\begin{equation}
f(x) = \sum_{q=1}^{2n+1} \Phi_q \left( \sum_{p=1}^{n} \phi_{q,p}(x_p) \right),
\end{equation} in which $x$ is an input vector containing $n$ variables, $x_{p}$ is the $p$-th element of $x$ and $\Phi_q$, $\phi_{q,p}$ are univariate functions corresponding to each variable.
This theorem shows that any high-dimensional continuous function $f$ can be broken down into a nested sum of one-dimensional functions. KAN extends this theory to create deep neural networks $\text{KAN}(x)$, with the overall computation represented as the composition of function matrices from each layer:
\begin{equation}
    \text{KAN}(x) = (\Phi_{L-1} \circ \Phi_{L-2} \circ \dots \circ \Phi_0)x,
\end{equation}where $x$ is the input vector, $\Phi_L$ is the function matrix of the $L$-th layer, and $\circ$ represents function composition. The fundamental innovation of KAN is moving the activation function $\phi{(x)}$ from fixed ``nodes" to learnable ``edges". The authors parameterize $\phi{(x)}$ on each connection using a B-spline:
\begin{equation}
   \phi(x) = w_b \text{SiLU}(x) + w_s \sum_{i} \omega_i B_i(x),
\end{equation}
where $\text{SiLU}(x)$ is the basis function, $B_i(x)$ is the B-spline basis function, $\omega_i$ are the learnable spline coefficients, and $w_{b}$ and $w_{s}$ are weights that control the relative size of the two components. This structure of placing functions on the edges and performing summations at the nodes entirely replaces the linear weight matrices of MLP. As a result, KAN can achieve greater approximation accuracy and offer strong interpretability.

\begin{table*}[t]	
	\centering\renewcommand\arraystretch{1.2}\setlength{\tabcolsep}{12.1pt}
	\belowrulesep=0pt\aboverulesep=0pt
	\caption{Cross-dataset evaluation of representative face forgery detection methods on CDF2, DFDC, DFDCP and FFIW datasets. Results for baseline methods are taken from their respective papers. The best performance is highlighted in \textbf{BOLD}, and the second best is \underline{underlined}. \label{cross-data}}
	\begin{tabular}{c|c|c|cc|cccc}
		\toprule
		\multirow{2}{*}{Method} & 
		\multirow{2}{*}{Venue} & 
		\multirow{2}{*}{Input Type} &
		\multicolumn{2}{c|}{Training Set} & \multicolumn{4}{c}{Test Set AUC (\%)}\\
		\cmidrule(lr){4-5}\cmidrule(lr){6-9}
		&\multicolumn{1}{c|}{} 
		&\multicolumn{1}{c|}{} 
		&\multicolumn{1}{c}{Real} 
		&\multicolumn{1}{c|}{Fake} 
		&\multicolumn{1}{c}{CDF2} 
		&\multicolumn{1}{c}{DFDC} 
		&\multicolumn{1}{c}{DFDCP}
		&\multicolumn{1}{c}{FFIW} \\
		\midrule
		SBI \cite{Shiohara_2022_CVPR} & CVPR 2022 & Frame & \usym{2713} &  & 93.18 & 72.42 & 86.15 & {84.83} \\
		SeeABLE \cite{Larue_2023_ICCV} & ICCV 2023 & Frame & \usym{2713} &  & 87.30 & 75.90 & 86.30 & - \\
		AUNet \cite{Bai_2023_CVPR} & CVPR 2023 & Frame & \usym{2713} &  & 92.77 & 73.82 & 86.16 & 81.45 \\
		LAA-Net \cite{Nguyen_2024_CVPR} & CVPR 2024 & Frame & \usym{2713} &  & 95.40 & - & 86.94 & - \\
		RAE \cite{10.1007/978-3-031-72943-0_23} & ECCV 2024 & Frame & \usym{2713} &  & \underline{95.50} & {80.20} & \underline{89.50} & - \\
		FreqBlender \cite{zhou2024freqblender} & NeurIPS 2024 & Frame & \usym{2713} &  & 94.59 & 74.59 & 87.56 & \underline{86.14} \\
		UDD \cite{fu2025exploring} & AAAI 2025 & Frame & \usym{2713} & \usym{2713} & 93.10 & {81.20} & 88.10 & - \\
		\midrule
		TALL++ \cite{xu2024towards} & IJCV 2024 & Video & \usym{2713} & \usym{2713} & 91.96 & - & 78.51 & - \\
		NACO \cite{zhang2025learning} & ECCV 2024 & Video & \usym{2713} & \usym{2713} & 89.50 & - & 76.70 & - \\
        DFD-FCG \cite{11093395} & CVPR 2025 & Video & \usym{2713} & \usym{2713} & 95.00 & \underline{81.81} & - & - \\    
		\midrule
		LAKAN (Ours) & - & Frame & \usym{2713} &  & \textbf{96.63} & \textbf{84.52} & \textbf{89.71} & \textbf{87.32} \\      
		\bottomrule
	\end{tabular}
    \vspace{-5mm}
\end{table*}

\begin{table}[t]	
	\centering\renewcommand\arraystretch{1.2}\setlength{\tabcolsep}{7pt}
	\belowrulesep=0pt\aboverulesep=0pt
	\caption{Cross-manipulation evaluation of methods trained only on real face videos from FF++. \label{cross-mani}}
	\begin{tabular}{c|cccc|c}
		\toprule
		\multirow{2}{*}{Method} & 
		\multicolumn{5}{c}{Test Set AUC (\%)} \\
		\cmidrule(lr){2-6}
		&\multicolumn{1}{c}{DF} 
		&\multicolumn{1}{c}{F2F} 
		&\multicolumn{1}{c}{FS}
		&\multicolumn{1}{c|}{NT}
		&\multicolumn{1}{c}{FF++} \\
		\midrule
		SBI \cite{Shiohara_2022_CVPR} & \underline{99.99} & \underline{99.88} & \textbf{99.91} & \underline{98.79} & \underline{99.64} \\
		SeeABLE \cite{Larue_2023_ICCV} & 99.20 & 98.80 & 99.10 & 96.90 & 98.50 \\
		AUNet \cite{Bai_2023_CVPR} & 99.98 & 99.60 & \underline{99.89} & 98.38 & 99.46 \\
		RAE \cite{10.1007/978-3-031-72943-0_23} & 99.60 & 99.10 & 99.20 & 97.60 & 98.90 \\
		\midrule
		LAKAN & \textbf{100} & \textbf{100 
} & 99.85 & \textbf{98.99} & \textbf{99.71}\\      
		\bottomrule
	\end{tabular}
    \vspace{-2.5mm}
\end{table}

\subsection{Architecture Overview}
Our overall architecture is based on an image encoder and integrates the Landmark-assisted Adaptive Kolmogorov-Arnold Network (LAKAN) module innovatively. Specifically, we insert LAKAN modules at the entrance of the main stages of image encoder before the first block of each stage, as illustrated in Fig. \ref{fig:figure2}. LAKAN utilizes facial landmark information detected using the Dlib library \cite{10.5555/1577069.1755843} to dynamically adjust feature weights at multiple scales, enabling the model to automatically focus on key facial regions to accomplish precise face forgery detection. Notably, this design allows LAKAN to function as a plug-and-play module, making it compatible with various image encoder architectures.

\subsection{The LAKAN Module}
As shown in Fig. \ref{fig:figure2}, the core design of LAKAN module is to utilize facial landmarks as structural prior information to generate a dynamic guidance signal for the image encoder that is adapted to the input instance. The module is centered around a dynamic KAN layer, whose internal parameters are not fixed during training but are generated in real-time based on the facial landmarks $L \in{\mathbb{R}}^{N_{lm} \times 2}$ of each input image. $N_{lm}$ is the number of landmarks. The dynamic parameter generation process is key to achieving adaptivity: it begins with applying sinusoidal positional embedding (PosEmbed) to landmark coordinates, followed by using a lightweight MLP network (MLP) to turn encoded high-dimensional features into a unified guidance vector $v_{guide}$. This vector encapsulates facial geometric information and is ultimately fed into two separate fully connected (FC) heads to produce two core parameter sets required by the dynamic KAN layer: the spline weight $W_{spline}$ and the spline scaler $S_{spline}$.
When a feature map $X$ is passed by the image encoder to LAKAN module, these parameters dynamically generated from the landmarks are then used to configure the behavior of the dynamic KAN layer in real time. The KAN layer processes the feature map $X$, and its non-linear transformation is determined by the input landmark structure, producing the output $X'$, The entire process can be summarized as follows:
\begin{equation}
\begin{aligned}
    X' &= \text{LAKAN}(X,L),\\
    \text{LAKAN}(X,L) &= \text{KAN}(X|W_{spline}, S_{spline}),\\
    W_{spline}, S_{spline} &= \text{FC}_{W}(v_{guide}), \text{FC}_{S}(v_{guide}),\\
    v_{guide} &= \text{MLP}(\text{PosEmbed}(L)).
\end{aligned}
\end{equation}
$X'$ is then passed through a sigmoid function $\sigma(\cdot)$ for normalization to generate the gating signal $G = \sigma(X')$. Finally, we apply this gating signal back to the feature map in a multiplicative form with a connection to obtain the guided output $X_{out} = X \odot (1 + G)$. Through this entire mechanism, the LAKAN module is able to dynamically enhance or suppress feature responses based on the unique structure of each face, thereby guiding the entire model to focus its attention on the key regions most likely to contain forgery artifacts.

\section{EXPERIMENTS}
\label{sec:experiments}
\subsection{Settings}
\textbf{Datasets.}
We use the FaceForensics++ (FF++) \cite{Rossler_2019_ICCV} dataset. According to SBI framework \cite{Shiohara_2022_CVPR}, our model is trained using 1000 authentic videos. The performance of the trained model is then evaluated on four types of forged videos provided by FF++: DeepFakes (DF), Face2Face (F2F), FaceSwap (FS), and NeuralTextures (NT). To evaluate the model's cross-dataset generalization capability, we adopt four public datasets with different challenges: Celeb-DeepFake-v2 (CDF2) \cite{Li_2020_CVPR}, which is characterized by high-quality forged videos of celebrities, DeepFake Detection Challenge (DFDC) \cite{dolhansky2020deepfake} and its preview version (DFDCP) \cite{dolhansky2019dee}, which contain extensive video perturbations and FFIW-10K (FFIW) \cite{Zhou_2021_CVPR}, which increases detection difficulty by including multi-person scenarios.

\noindent
\textbf{Baselines.}
We compare our method with 10 representative frame-level and video-level detection baselines, including SBI \cite{Shiohara_2022_CVPR}, SeeABLE \cite{Larue_2023_ICCV}, AUNet \cite{Bai_2023_CVPR}, LAA-Net \cite{Nguyen_2024_CVPR}, RAE \cite{10.1007/978-3-031-72943-0_23}, FreqBlender \cite{zhou2024freqblender}, UDD \cite{fu2025exploring}, TALL++ \cite{xu2024towards}, NACO \cite{zhang2025learning} and DFD-FCG \cite{11093395}. 

\noindent
\textbf{Evaluation Metric.}
To evaluate and compare the detection performance of different methods, we adopt the Area Under the Receiver Operating Characteristic Curve (AUC) as the primary metric, which is a standard metric in the field of face forgery detection. To ensure a fair and direct comparison with video-level approaches, we convert our frame-level predictions to video-level results by averaging them across all frames of each video.

\noindent
\textbf{Implementation Details.}
Our experimental setup is configured as follows. We strictly follow the synthetic data generation, preprocessing, and augmentation strategies from SBI \cite{Shiohara_2022_CVPR}. Our image encoder is a ConvNeXt-Base network \cite{9879745} which consists of 3, 3, 27 and 3 blocks across its four stages. The number of landmarks $N_{lm}$ per image is 68. It is initialized with weights pre-trained on ImageNet-1K dataset. The model is trained for 200 epochs using the AdamW optimizer, a batch size of 64, and an initial learning rate of 5e-5, with a linear learning rate decay scheduled after the 100th epoch. Our implementation is based on PyTorch and requires approximately 23 GB of GPU memory for training.

\begin{table}[t]	
    \centering\renewcommand\arraystretch{1.2}\setlength{\tabcolsep}{7pt}
	\belowrulesep=0pt\aboverulesep=0pt
	\caption{Ablation study results for LAKAN. \label{LAKAN_ablation}}
    \begin{tabular}{cc|cccc}
    \toprule
    \multicolumn{2}{c|}{Method} & \multicolumn{4}{c}{Test Set AUC (\%)} \\ \cmidrule(lr){1-2} \cmidrule(lr){3-6} 
    KAN &landmarks & CDF2 & DFDC & DFDCP & FFIW   \\ 
    \midrule
     \usym{2717} & \usym{2717} & \underline{96.19} & 82.01 & 86.84 & \underline{86.36}\\
    \usym{2717}  & \usym{2713} & 95.59 & 83.28 & \underline{88.51} & 83.17 \\
    \usym{2713} & \usym{2717} & 95.24 & \underline{84.13} & 88.34 & 85.75
      \\ \midrule
\usym{2713} & \usym{2713} &  \textbf{96.63} & \textbf{84.52}               & \textbf{89.71}   & \textbf{87.32}   \\ \bottomrule
\end{tabular}

\vspace{-1.5mm}

\centering\renewcommand\arraystretch{1.2}\setlength{\tabcolsep}{9pt}
	\belowrulesep=0pt\aboverulesep=0pt
	\caption{Ablation study results for fusion strategy. \label{Fusion_ablation}}
	\begin{tabular}{c|cccc}
		\toprule
		\multirow{2}{*}{Method} & 
		\multicolumn{4}{c}{Test Set AUC (\%)} \\
		\cmidrule(lr){2-5}
		&\multicolumn{1}{c}{CDF2} 
		&\multicolumn{1}{c}{DFDC} 
		&\multicolumn{1}{c}{DFDCP}
		&\multicolumn{1}{c}{FFIW} \\
		\midrule
		 Addition  & \underline{95.97} & \underline{83.08} & \underline{88.15} & \underline{86.22} \\
		 Product  & 91.22 & 76.91 & 84.20 & 73.64 \\
		 Concatenation  & 87.64 & 77.04 & 77.92 & 69.57 \\
		\midrule
		Gating & \textbf{96.63} & \textbf{84.52} & \textbf{89.71} & \textbf{87.32}\\ 
		\bottomrule
	\end{tabular}

\vspace{-1.5mm}

\centering\renewcommand\arraystretch{1.2}\setlength{\tabcolsep}{5.7pt}
	\belowrulesep=0pt\aboverulesep=0pt
	\caption{Ablation study results for different image encoders. \label{backbone_ablation}}
	\begin{tabular}{c|cccc|c}
		\toprule
		\multirow{2}{*}{Backbone} & 
		\multicolumn{5}{c}{Test Set AUC (\%)} \\
		\cmidrule(lr){2-6}
		&\multicolumn{1}{c}{CDF2} 
		&\multicolumn{1}{c}{DFDC} 
		&\multicolumn{1}{c}{DFDCP}
		&\multicolumn{1}{c|}{FFIW} 
		&\multicolumn{1}{c}{Avg.} \\
		\midrule
		EfficientNet-B4  & 93.18 & 72.42 & 86.15 & 84.83 & 84.15 \\
		+ LAKAN  & 94.57 & 80.09 & 87.54 & 85.02 & 86.81 \\ \midrule
		Swin-B  & 95.53 & 82.81 & 89.70 & {85.38} & 88.36 \\
		+ LAKAN  & 96.00 & \underline{84.17} & \textbf{91.08} & 85.78 & \underline{89.26} \\ \midrule
		ConvNeXt-B  & \underline{96.19} & 82.01 & 86.84 & \underline{86.36} & {87.85} \\
		+ LAKAN  & \textbf{96.63} & \textbf{84.52} & \underline{89.71} & \textbf{87.32} & \textbf{89.55} \\
		\bottomrule
	\end{tabular}
    \vspace{-5mm}
    
\end{table}

\subsection{Results}
\noindent
\textbf{Cross-Dataset Evaluation.}
We conduct a cross-dataset evaluation, a vital protocol for testing generalization. As detailed in Table \ref{cross-data}, our method is compared against recent SOTA models. Its superior generalization stems from our LAKAN module, which guides the model to focus on fundamental forgery patterns across datasets. This allows it to consistently achieve top AUC scores on unseen datasets.

\noindent
\textbf{Cross-Manipulation Evaluation.}
In practice, defenders rarely know the specific forgery techniques used by attackers, making model's ability to generalize across different manipulation types essential. Following the protocol in \cite{Shiohara_2022_CVPR}, we compare our method against others trained exclusively on real faces from FF++. The results in Table \ref{cross-mani} show that our model with LAKAN is effective at detecting unseen manipulation techniques. This effectiveness is attributed to LAKAN's landmark-based guidance, which focuses on structural inconsistencies to most forgery methods, instead of specific manipulation artifacts.

\begin{figure}[htb]
  \centering
  \includegraphics[width=8cm]{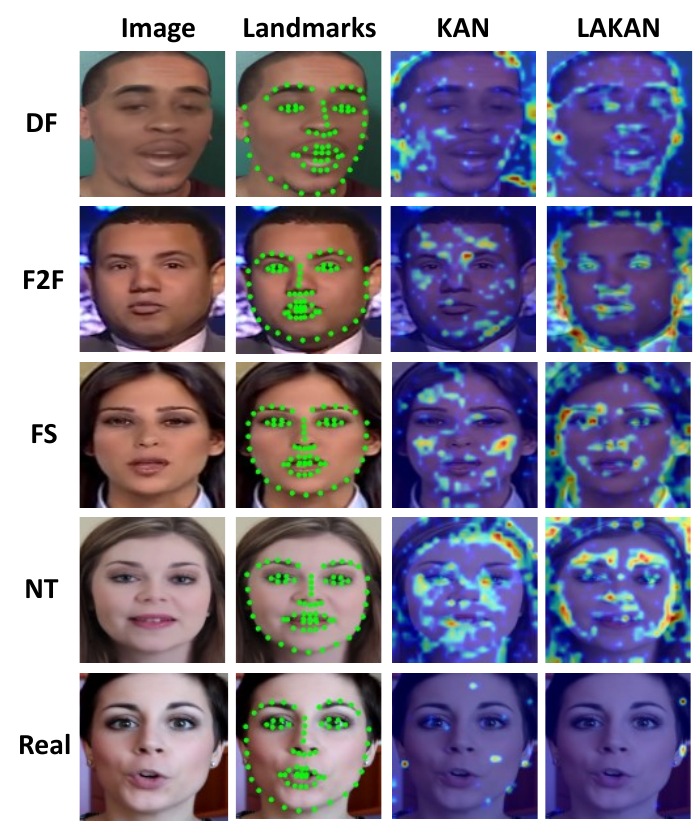}
  \vspace{-3mm}
  \caption{Landmarks and saliency maps on four distinct manipulated samples and one real sample from the FF++ dataset.}
  \label{fig:figure3}
  \vspace{-2mm}
\end{figure}

\subsection{Ablation Studies}
\noindent
\textbf{Analysis of LAKAN.}
We conduct ablation studies on KAN and landmark information to validate our design, presented in Table \ref{LAKAN_ablation}. Although a single component's bias can slightly reduce performance on certain datasets, the full LAKAN model consistently achieves superior performance, proving its adaptive mechanism combines both components to overcome individual limitations.

\noindent
\textbf{Effectiveness of Gating Mechanism.}
We investigate effectiveness of the gating mechanism versus alternative feature fusion strategies (Addition, Product, Concatenation). The cross-dataset evaluation results, presented in Table \ref{Fusion_ablation}, show the superiority of gating mechanism.

\noindent
\textbf{Analysis of Different Image Encoders.}
LAKAN is a plug-and-play module and we have conducted experiments on different image encoders, including EfficientNet \cite{tan2019efficientnet} and Swin Transformer\cite{Liu_2021_ICCV}. Similar to the ConvNeXt-Base image encoder, we insert LAKAN after each feature downsampling stage. As shown in Table \ref{backbone_ablation}, the LAKAN module brings significant performance improvements to all tested image encoders, proving its general applicability. The average performance on the four datasets shows that the ConvNeXt-based LAKAN model performs the best. Therefore, our experiments are mainly carried out on the ConvNeXt image encoder.

\noindent
\textbf{Visual Analysis of LAKAN.}
We further validate LAKAN's effectiveness through Grad-CAM visualizations, highlighting key decision regions. As shown in the saliency maps in Fig. \ref{fig:figure3}, for fake samples, LAKAN directs the model’s attention toward key facial contours rich in forgery artifacts. Conversely, the model exhibits minimal activation for the real sample, indicating it does not focus on any specific area when an image is authentic.

\section{CONCLUSION}
In this paper, we propose LAKAN, a novel plug-and-play module to enhance the generalization of face forgery detectors. Our core contribution is a novel fusion strategy that integrates the facial landmark priors with deep visual features through a dynamically parameterized Kolmogorov-Arnold Network (KAN). This allows the model to perform adaptive feature modulation, focusing on the semantically significant facial regions. Extensive experiments show that LAKAN achieves superior performance on challenging cross-dataset and cross-manipulation benchmarks. Its consistent performance gains across various modern encoders prove its effectiveness in handling unseen forgery techniques.

\vfill\pagebreak

\apptocmd{\thebibliography}{\setlength{\itemsep}{0pt}}{}{}
\bibliographystyle{IEEEbib}
\bibliography{strings,refs}

\end{document}